\documentclass[runningheads, envcountsame, a4paper]{llncs}
\usepackage{graphicx}
\usepackage{times}
\usepackage{epsfig}
\usepackage{graphicx}
\usepackage{amsmath}
\usepackage{amssymb}
\usepackage{amsmath}
\usepackage{multirow}
\usepackage{graphicx}
\usepackage{booktabs} 
\usepackage{amsfonts}
\usepackage{subfigure}
\usepackage[misc]{ifsym}
\usepackage[table]{xcolor}
\def\bs{\boldsymbol}
\begin{document}
\title{Dep-$L_0$: Improving $L_0$-based Network Sparsification via  Dependency Modeling}
%
%
\author{Yang Li, \and
Shihao Ji~\Letter}
\authorrunning{Y. Li, and S. Ji}
%
\institute{Georgia State University, GA, USA\\
\email{yli93@student.gsu.edu, sji@gsu.edu}}
\tocauthor{Yang~Li, and Shihao~Ji}
\toctitle{Dep-$L_0$: Improving $L_0$-based Network Sparsification via  Dependency Modeling}
\maketitle              
\begin{abstract}
    Training deep neural networks with an $L_0$ regularization is one of the prominent approaches for network pruning or sparsification. The method prunes the network during training by encouraging weights to become exactly zero. However, recent work of Gale et al.~\cite{gale2019state} reveals that although this method yields high compression rates on smaller datasets, it performs inconsistently on large-scale learning tasks, such as ResNet50 on ImageNet. We analyze this phenomenon through the lens of variational inference and find that it is likely due to the independent modeling of binary gates, the mean-field approximation~\cite{viBlei}, which is known in Bayesian statistics for its poor performance due to the crude approximation. To mitigate this deficiency, we propose a dependency modeling of binary gates, which can be modeled effectively as a multi-layer perceptron (MLP). We term our algorithm Dep-$L_0$ as it prunes networks via a dependency-enabled $L_0$ regularization. Extensive experiments on CIFAR10, CIFAR100 and ImageNet with VGG16, ResNet50, ResNet56 show that our Dep-$L_0$ outperforms the original $L_0$-HC algorithm of Louizos et al.~\cite{louizos2017learning} by a significant margin, especially on ImageNet. Compared with the state-of-the-arts network sparsification algorithms, our dependency modeling makes the $L_0$-based sparsification once again very competitive on large-scale learning tasks. Our source code is available at \url{https://github.com/leo-yangli/dep-l0}.

\end{abstract}
%
%
%

\section{Introduction} 
Convolutional Neural Networks (CNNs) have achieved great success in a broad range of tasks. However, the huge model size and high computational price make the deployment of the state-of-the-art CNNs to resource-limited embedded systems (e.g., smart phones, drones and surveillance cameras) impractical. To alleviate this problem, substantial efforts have been made to compress and speed up the networks~\cite{model_compression}. Among these efforts, network pruning has been proved to be an effective way to compress the model and speed up inference without losing noticeable accuracy~\cite{lecun1990optimal,han2015learning,wen2016ssl,louizos2017learning,li2019l_0,you2019gate,gsm19,park2020lookahead}.

\begin{figure}[ht]
  \begin{center}
    \includegraphics[width=0.7\linewidth]{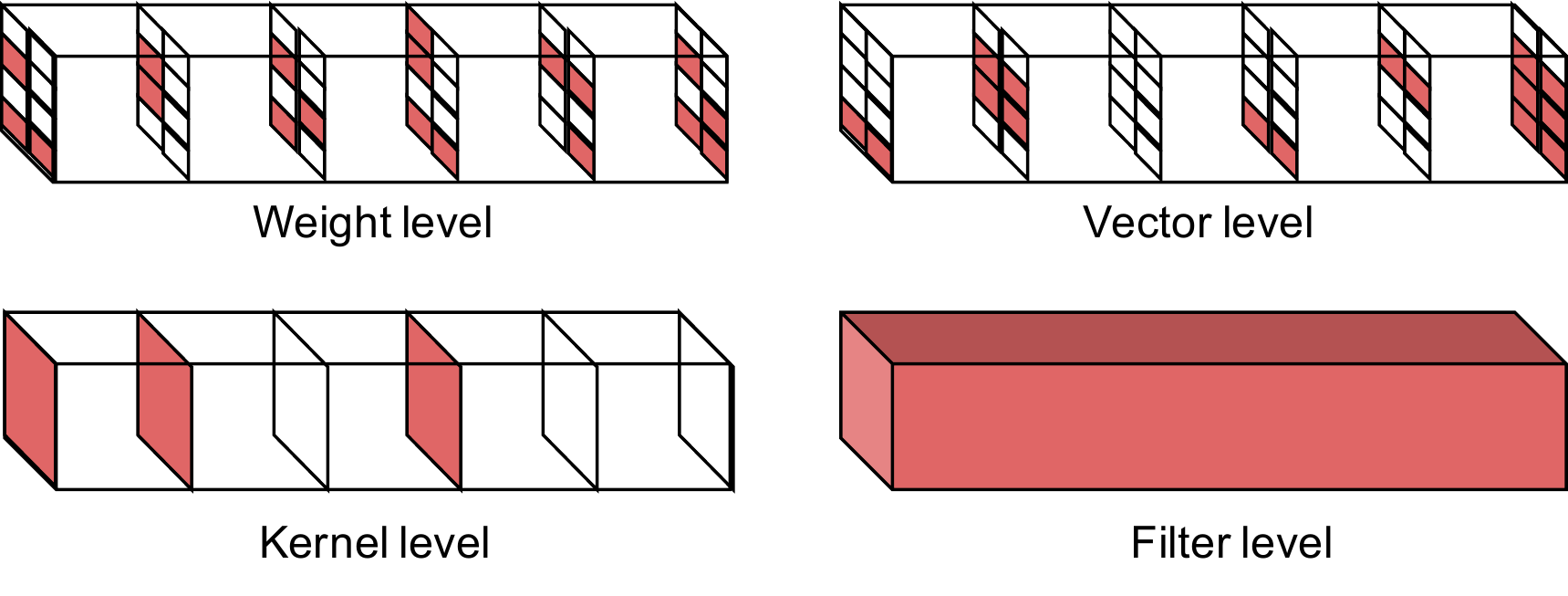}
  \end{center} \vspace{-20pt}
  \caption{Visualization of the weights of a convolutional filter and different pruning granularities. The red regions highlight the weights that can be pruned by different pruning methods. This paper focuses on filter-level pruning.}\label{fig:gran}
\end{figure}

The existing network pruning algorithms can be roughly categorized into two categories according to the pruning granularity: unstructured pruning~\cite{lecun1990optimal,han2015learning,gsm19,park2020lookahead} and structured pruning~\cite{li2016pruning,wen2016ssl,liu2017learning,zhuang2018discrimination,ding2019approximated,you2019gate,lin2020hrank}. As shown in Fig.~\ref{fig:gran}, unstructured pruning includes weight-level, vector-level and kernel-level pruning, while structured pruning normally refers to filter-level pruning. Although unstructured pruning methods usually lead to higher prune rates than structured ones, they require specialized hardware or software to fully utilize the benefits induced by the high prune rates due to the irregular network structures yielded by unstructured pruning. On the other hand, structured pruning can maintain the regularity of network structures, while pruning the networks effectively, and hence can fully utilize the parallel computing resources of general-purpose CPUs or GPUs. Because of this, in recent years structured pruning has attracted a lot of attention and achieved impressive performances~\cite{liu2017learning,zhuang2018discrimination,ding2019approximated,lin2020hrank}. In this work, we focus on structured pruning, more specifically, filter-level pruning. 

In terms of pruning methods, a simple yet effective strategy is heuristic-based, e.g., pruning the weights based on their magnitudes~\cite{lecun1990optimal,han2015learning,li2016pruning,hu2016network}. Another popular approach is to penalize the model size via sparsity inducing regularization, such as $L_1$ or $L_0$ regularization~\cite{liu2015sparse,wen2016ssl,louizos2017learning}. Among them, $L_0$-HC~\cite{louizos2017learning} is one of the state-of-the-art pruning algorithms that incorporates $L_0$ regularization for network pruning and has demonstrated impressive performances on many image classification benchmarks (e.g., MNIST, CIFAR10 and CIFAR100). This method attaches a binary gate to each weight of a neural network, and penalizes the complexity of the network, measured by the $L_0$ norm of the weight matrix. However, recent work of Gale et al.~\cite{gale2019state} reveals that although $L_0$-HC works well on smaller datasets, it fails to prune very deep networks on large-scale datasets, such as ResNet50 on ImageNet. The original $L_0$-HC algorithm was proposed and evaluated on filter-level pruning, while Gale et al.~\cite{gale2019state} focus on the weight-level pruning. Therefore, it is unclear if the observation of~\cite{gale2019state} is due to pruning granularity or the deficiency of the $L_0$ regularization based method. To understand this, we evaluate the original $L_0$-HC to sparsify ResNet50 at filter level on ImageNet, and find that it indeed cannot prune ResNet50 without a significant damage of model quality, confirming the observation made by~\cite{gale2019state}. This indicates that the failure of $L_0$-HC is likely due to the deficiency of the $L_0$-norm based approach. We further analyze $L_0$-HC in the lens of variational inference~\cite{viBlei}, and find that \textbf{the failure is likely due to an over-simplified assumption that models the variational posterior of binary gates to be element-wise independent}. To verify this hypothesis, we propose to incorporate the dependency into the binary gates, and model the gate dependency across CNN layers with a multi-layer perceptron (MLP). Extensive experiments show that our dependency-enabled $L_0$ sparsification, termed Dep-$L_0$, once again is able to prune very deep networks on large-scale datasets, while achieving competitive or sometimes even better performances than the state-of-the-art pruning methods. 

Our main contributions can be summarized as follows:
\begin{itemize}
\item From a variational inference perspective, we show that the effectiveness of $L_0$-HC~\cite{louizos2017learning} might be hindered by the implicit assumption that all binary gates attached to a neural network are independent to each other. To mitigate this issue, we propose Dep-$L_0$ that incorporates the dependency into the binary gates to improve the original $L_0$-based sparsification method.

\item A series of experiments on multiple datasets and multiple modern CNN architectures demonstrate that Dep-$L_0$ improves $L_0$-HC consistently, and is very competitive or sometimes even outperforms the state-of-the-art pruning algorithms.

\item Moreover, Dep-$L_0$ converges faster than $L_0$-HC in terms of network structure search, and reduces the time to solution by 20\%-40\% compared to $L_0$-HC in our experiments. 
\end{itemize}

\section{Related Work}

Model compression~\cite{model_compression} aims to reduce the size of a model and speed up its inference at the same time. Recently, there has been a flurry of interest in model compression, ranging from network pruning~\cite{lecun1990optimal,han2015learning,louizos2017learning,li2019l_0,lin2019toward}, quantization and binarization~\cite{courbariaux2016binarized,gupta2015deep}, tensor decomposition~\cite{jaderberg2014speeding,denton2014exploiting}, and knowledge distillation~\cite{hinton2015distilling}. Since our algorithm belongs to the category of network pruning, we mainly focus on reviewing related work in pruning.

\paragraph{\textbf{Network Pruning}}
A large subset of pruning methods is heuristic-based, which assigns an importance score to each weight and prune the weights whose importance scores are below a threshold. The importance scores are usually devised according to types of networks, e.g., the magnitude of weights~\cite{lecun1990optimal,han2015learning} (Feed-forward NNs), the $L_1$ or $L_2$ norm of filters~\cite{li2016pruning} (CNNs), and the average percentage of zero activations~\cite{hu2016network} (CNNs). However, Ye et al.~\cite{ye2018rethinking} point out that the assumption that weights/filters of smaller norms are less important may not hold in general, challenging the heuristic-based approaches. These methods usually follow a three-step training procedure: training - pruning - retraining in order to achieve the best performance.

Another subset of pruning methods focuses on training networks with sparsity inducing regularizations. For example, $L_2$ and $L_1$ regularizations~\cite{liu2015sparse,wen2016ssl} or $L_0$ regularization~\cite{louizos2017learning} can be incorporated into the objective functions to train sparse networks. Similarly, Molchanov et al.~\cite{molchanov2017variational} propose variational dropout, a sparse Bayesian learning algorithm under an improper logscale uniform prior, to induce sparsity. In this framework, network pruning can be performed from scratch and gradually fulfilled during training without separated training stages.

Recently, Gale et al.~\cite{gale2019state} evaluate three popular pruning methods, including variational dropout~\cite{molchanov2017variational}, $L_0$-HC~\cite{louizos2017learning} and magnitude-based pruning~\cite{zhu2017prune}, on two large-scale benchmarks. They reveal that although $L_0$-HC~\cite{louizos2017learning} is more sophisticated and yields state-of-the-art results on smaller datasets, it performs inconsistently on large-scale learning task of ImageNet. This observation motivates our development of Dep-$L_0$. To mitigate the deficiency of $L_0$-HC, we propose dependency modeling, which makes the $L_0$-based pruning once again competitive on large-scale learning tasks.

\paragraph{\textbf{Dependency Modelling}}
Even though there are many network pruning algorithms today, most of them (if not all) \emph{implicitly} assume all the neurons of a network are independent to each other when selecting neurons for pruning. There are quite few works exploring the dependency inside neural networks for pruning. The closest one is LookAhead~\cite{park2020lookahead}, which reinterprets the magnitude-based pruning as an optimization of the Frobenius distortion of a single layer, and improves the magnitude-based pruning by optimizing the Frobenius distortion of multiple layers, considering previous layer and next layer. Although the interaction of different layers is considered, the authors do not model the dependency of them explicitly. To the best of our knowledge, our Dep-$L_0$ is the first to model the dependency of neurons explicitly for network pruning.

\section{Method}

Our algorithm is motivated by $L_0$-HC~\cite{louizos2017learning}, which prunes neural networks by optimizing an $L_0$ regularized loss function and relaxing the non-differentiable Bernoulli distribution with the Hard Concrete (HC) distribution. 
Since $L_0$-HC can be viewed as a special case of variational inference under the spike-and-slab prior ~\cite{louizos2017learning}, in this section we first formulate the sparse structure learning from this perspective, then discuss the deficiency of $L_0$-HC and propose dependency modeling, and finally present Dep-$L_0$.

\subsection{Sparse Structure Learning}


Consider a dataset $\mathcal{D}=\left\{\bs{x}_{i}, y_{i}\right\}_{i=1}^{N}$ that consists of $N$ pairs of instances, where $\bs{x}_i$ is the $i$th observed data and $y_i$ is the associated class label. We aim to learn a model $p(\mathcal{D}|\bs \theta)$, parameterized by $\bs{\theta}$, which fits $\mathcal{D}$ well with the goal of achieving good generalization to unseen test data. In order to sparsify the model, we introduce a set of binary gates $\bs{z}=\{z_1,\cdots,z_{|\bs{\theta}|}\}$, one gate for each parameter, to indicate whether the corresponding parameter being kept $(z=1)$ or not $(z=0)$. 


This formulation is closely related to the spike-and-slab distribution~\cite{mitchell1988bayesian}, which is widely used as a prior in Bayesian inference to impose sparsity. Specifically, the spike-and-slab distribution defines a mixture of a delta spike at zero and a standard Gaussian distribution:
\begin{align}
  &\quad\quad\quad\quad\quad p(z)=\text{Bern}(z|\pi)\nonumber\\
  &p(\theta|z=0)=\delta(\theta), \quad p(\theta|z=1)=\mathcal{N}(\theta|0, 1),
\end{align}
where $\text{Bern}(\cdot|\pi)$ is the Bernoulli distribution with parameter $\pi$, $\delta(\cdot)$ is the Dirac delta function, i.e., a point probability mass centered at the origin, and $\mathcal{N}(\theta|0, 1)$ is the Gaussian distribution with zero mean and unit variance. Since both $\bs{\theta}$ and $\bs{z}$ are vectors, we assume the prior $p(\bs{\theta},\bs{z})$ factorizes over the dimensionality of $\bs{z}$.

In Bayesian statistics, we would like to estimate the posterior of $(\bs{\theta}, \bs{z})$, which can be calculated by Bayes' rule:
\begin{equation}
  p(\bs\theta, \bs{z} | \mathcal{D}) = \frac{p(\mathcal{D}|\bs{\theta},\bs{z}) p(\bs{\theta},\bs{z})}{p(\mathcal{D})}.
\end{equation}
Practically, the true posterior distribution $p(\bs{\theta},\bs{z}|\mathcal{D})$ is intractable due to the non-conjugacy of the model likelihood and the prior. Therefore, here we approximate the posterior distribution via variational inference~\cite{viBlei}. Specially, we can approximate the true posterior with a parametric variational posterior $q(\bs{\theta},\bs{z})$, the quality of which can be measured by the Kullback-Leibler (KL) divergence: 
\begin{equation} \label{eq: KL}
  KL[q (\bs{\theta}, \bs{z}) || p(\bs{\theta}, \bs{z} | \mathcal{D})],
\end{equation}
which is again intractable, but can be optimized by maximizing the variational lower bound of $\log p(\mathcal{D})$, defined as
\begin{equation} \label{eq: lowerbound}
  L= \mathbb{E}_{q(\bs{\theta}, \bs{z})}[\log p(\mathcal{D} | \bs{\theta}, \bs{z})] - KL[q(\bs{\theta},\bs{z}) \| p(\bs{\theta},\bs{z})],
\end{equation}
where the second term can be further expanded as:
  \begin{align} \label{2term}
  KL[q(\bs{\theta}, \bs{z}) \| p(\bs{\theta}, \bs{z})]
  = &\mathbb{E}_{q(\bs{\theta}, \bs{z})} [\log{q (\bs{\theta}, \bs{z})} - \log{p(\bs{\theta},\bs{z})}]\nonumber\\
    = &\mathbb{E}_{q(\bs{\theta},\bs{z})} [\log{q (\bs{\theta} | \bs{z})} - \log{p(\bs{\theta} | \bs{z})} + \log{q(\bs{z})-\log p(\bs{z})}]\nonumber\\
    = & KL[q(\bs{\theta} | \bs{z}) || p(\bs{\theta} | \bs{z})] + KL[q(\bs{z}) || p(\bs{z})].
  \end{align}
 In $L_0$-HC~\cite{louizos2017learning}, the variational posterior $q(\bs{z})$ is factorized over the dimensionality of $\bs{z}$, i.e., $q(\bs{z})=\prod_{j=1}^{|\bs{\theta}|}q(z_j)=\prod_{j=1}^{|\bs{\theta}|}\text{Bern}(z_j|\pi_j)$. By the law of total probability, we can further expand Eq. \ref{2term} as
    \begin{align} \label{2term2}
    &KL[q(\bs{\theta}, \bs{z}) \| p(\bs{\theta}, \bs{z})]\nonumber \\
    & = \sum_{j=1}^{|\bs{\theta}|} \Big( q(z_j=0)KL[q(\theta_j | z_j=0) || p(\theta_j | z_j=0)] + \nonumber\\
    & \quad + q(z_j=1)KL[q(\theta_j | z_j=1) || p(\theta_j | z_j=1)]\Big) +\sum_{j=1}^{|\bs{\theta}|} KL[q(z_j)|| p(z_j)]\nonumber\\
    &=\sum_{j=1}^{|\bs{\theta}|} q(z_j=1)KL[q(\theta_j | z_j=1) || p(\theta_j | z_j=1)] + \sum_{j=1}^{|\bs{\theta}|} KL[(q(z_j)|| p(z_j)].
  \end{align}
The last step holds because $KL[q(\theta_j | z_j=0) || p(\theta_j|z_j=0)]=KL[q(\theta_j | z_j=0) || \delta(\theta_j)]=0$.

Furthermore, letting $\bs{\theta} = \tilde{\bs{\theta}} \odot \bs{z}$ and assuming $\lambda = KL[q(\theta_j | z_j=1) || p(\theta_j |z_j=1)]$, the lower bound $L$ (\ref{eq: lowerbound}) can be simplified as
\begin{align}
  L &= \mathbb{E}_{q(\bs{z})}[\log p(\mathcal{D} | \tilde{\bs{\theta}} \odot \bs{z})] - \sum_{j=1}^{|\bs{\theta}|} K L\left(q\left({z}_{j}\right)|| p\left({z}_{j}\right)\right)-\lambda \sum_{j=1}^{|\bs{\theta}|} q\left({z}_{j}=1\right) \nonumber\\
  & \leq\mathbb{E}_{q(\bs{z})}[\log p(\mathcal{D} | \tilde{\bs{\theta}} \odot \bs{z})]-\lambda \sum_{j=1}^{|\bs{\theta}|} \pi_{j}, \label{F}
\end{align}
where the inequality holds due to the non-negativity of KL-divergence.

Given that our model is a neural network $h(\bs x;\tilde{\bs{ \theta}}, \bs{z})$, parameterized by $\tilde{\bs{\theta}}$ and $\bs{z}$, Eq.~\ref{F} turns out to be an $L_0$-regularized loss function~\cite{louizos2017learning}:
\begin{equation} \label{risk}
  \mathcal{R}(\tilde{\bs{\theta}}, \bs{\pi})=\mathbb{E}_{q(\bs{z} )}\left[\frac{1}{N}\sum_{i=1}^{N} \mathcal{L}\left(h(\bs{x}_{i} ; \tilde{\bs{\theta}} \odot \bs{z}), y_i\right)\right]+\lambda \sum_{j=1}^{|\bs{\theta}|} \pi_{j},
\end{equation}
where $\mathcal{L}(\cdot)$ is the cross entropy loss for classification.

In the derivations above, the variational posterior $q(\bs{z})$ is assumed to factorize over the dimensionality of $\bs{z}$, i.e., $q(\bs{z})=\prod_{j=1}^{|\bs{\theta}|}q(z_j)$. This means all the binary gates $\bs{z}$ are assumed to be \textit{independent} to each other -- the mean-field approximation~\cite{viBlei}. In variational inference, it is common to assume the prior $p(\bs{z})$ to be element-wise independent; the true posterior $p(\bs{z}|\mathcal{D})$, however, is unlikely to be element-wise independent. Therefore, approximating the true posterior by an element-wise independent $q(\bs{z})$ is a very restrict constraint that limits the search space of admissible $q(\bs{z})$ and is known in Bayesian statistics for its poor performance~\cite{bishop2007,viBlei}. We thus hypothesize that this mean-field approximation may be the cause of the failure reported by Gale et al.~\cite{gale2019state}, and the independent assumption hinders the effectiveness of the $L_0$-based pruning method. Therefore, we can potentially improve $L_0$-HC by relaxing this over-simplified assumption and modeling the dependency among binary gates $\bs{z}$ explicitly. 

Specifically, instead of using a fully factorized variational posterior $q(\bs{z})$, we can model $q(\bs{z})$ as a conditional distribution by using the chain rule of probability
\[
q(\bs{z}) = q(z_1) q(z_2|z_1) q(z_3|z_1,z_2) \cdots q(z_{|\theta|} | z_1,\cdots, z_{|\bs{\theta}|-1}),
\]
where given an order of binary gates $\bs{z}=\{z_1,z_2,\cdots,z_{|\theta|}\}$, $z_i$ is dependent on all previous gates $z_{<i}$. With this, Eq.~\ref{risk} can be rewritten as
  \begin{equation}\label{risk2}
    \mathcal{R}(\tilde{\bs{\theta}}, \bs{\pi}) = \lambda \sum_{j=1}^{|\bs{\theta}|} \pi_{j}+ \mathbb{E}_{q(z_1)\cdots q(z_{|\bs{\theta}|}|z_1,\cdots,z_{|\bs{\theta}|-1})}\left[\frac{1}{N}\sum_{i=1}^{N} \mathcal{L}\left(h(\bs{x}_{i} ; \tilde{\bs{\theta}} \odot \bs{z}), y_i\right)\right],
  \end{equation}
which is a dependency-enabled $L_0$ regularized loss function for network pruning. Detailed design of the dependency modeling is to be discussed in later sections.

\subsection{Group Sparsity}\label{sec:group}
So far we have modeled a sparse network by attaching a set of binary gates $\bs{z}$ to the network at the weight level. As we discussed in the introduction, we prefer to prune the network at the filter level to fully utilize general purpose CPUs or GPUs. To this end, we consider group sparsity that shares a gate within a group of weights. Let $G=\{g_1,g_2,\cdots,g_{|G|}\}$ be a set of groups, where each element corresponds to a group of weights, and $|G|$ is the number of groups. With the group sparsity, the expected $L_0$-norm of model parameters (the first term of Eq.~\ref{risk2}) can be calculated as 
\begin{equation}
  \mathbb{E}_{q(\bs{z})}\|\bs{\theta}\|_{0}=\sum_{j=1}^{|\bs{\theta}|} q(z_j=1|z_{<j})=\sum_{k=1}^{|G|} |g_k|\pi_{k},
\end{equation}
where $|g_k|$ denotes the number of weights in group $k$.

In all our experiments, we perform filter-level pruning by attaching a binary gate to all the weights of a filter (i.e., a group). Since modern CNN architectures often contain batch normalization layers~\cite{batchnorm}, in our implementation we make a slight modification that instead of attaching the gates to filters directly, we attach the gates to the feature maps after batch normalization. This is because batch normalization accumulates a moving average of feature statistics for normalization during the training process. Simply attaching a binary gate to the weights of a filter cannot remove the impact of a filter completely when $z=0$ due to the memorized statistics from batch normalization. By attaching the gates to the feature maps after batch normalization, the impact of the corresponding filter can be completely removed when $z=0$.

\subsection{Gate Partition}
\begin{figure}[t]
  \begin{center}
    \subfigure[element-wise dependency]{\raisebox{7mm}{\includegraphics[width=0.45\linewidth]{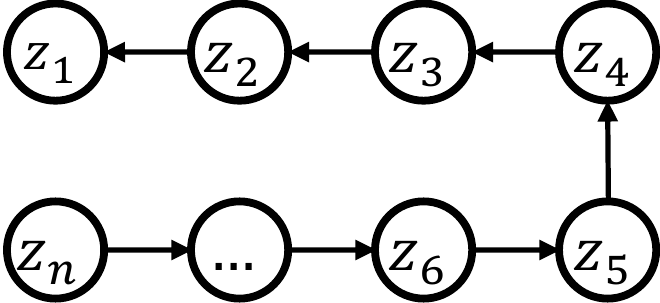}}}\hfill
    \subfigure[partition-wise dependency]{\includegraphics[width=0.45\linewidth]{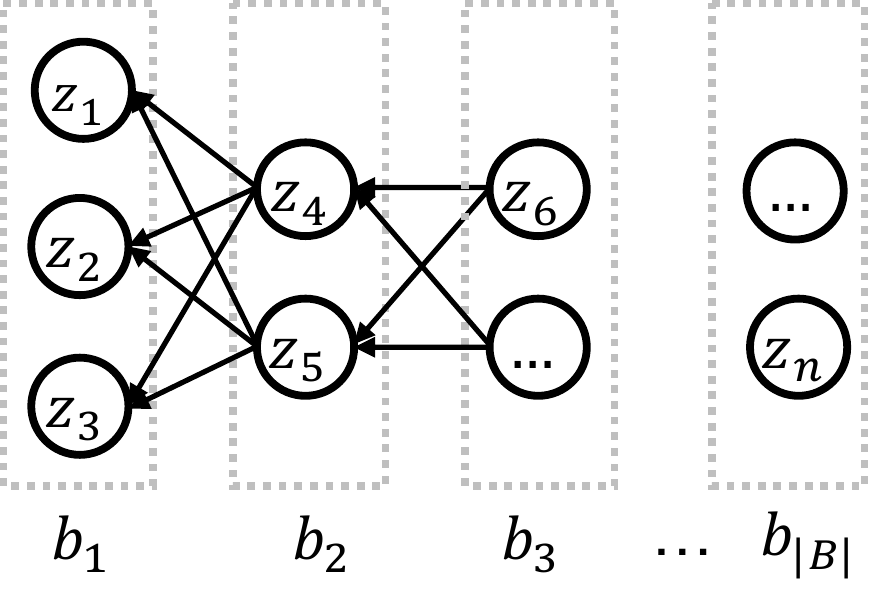}}
  \end{center}\vspace{-15pt}
  \caption{Illustration of (a) element-wise sequential dependency modeling, and (b) partition-wise dependency modeling.}\label{fig: gate partition}
\end{figure}

Modern CNN architectures, such as VGGNet~\cite{simonyan2014very}, ResNet~\cite{he2016deep} and WideResNet~\cite{wideresnet}, often come with a large number of weights and filters. For example, VGG16~\cite{simonyan2014very} contains 138M parameters and 4,224 filters. Since we attach a binary gate to each filter, the number of gates would be large and modeling the dependencies among them would lead to a huge computational overhead and optimization issues. To make our dependency modeling more practical, we propose gate partition to simplify the dependency modeling among gates. Specifically, the gates are divided into blocks, and the gates within each block are considered independent to each other, whereas the gates cross blocks are considered dependent. Fig.~\ref{fig: gate partition} illustrates the difference between an element-wise sequential dependency modeling and a partition-wise dependency modeling. Let's consider $z_1$, $z_2$, $z_3$ and $z_4$ in these two cases. In the element-wise sequential dependency modeling, as shown in Fig.~\ref{fig: gate partition}(a), $z_2$ is dependent on $z_1$, $z_3$ is dependent on $z_1$ and $z_2$, and so on. As number of gates could be very large, the element-wise sequential modeling would lead to a very long sequence, whose calculation would incur huge computational overhead. Instead, we can partition the gates, as in Fig.~\ref{fig: gate partition}(b), where $z_1$, $z_2$ and $z_3$ are in block $\bs{b}_1$ so they are considered independent to each other, while $z_4$ in block $\bs{b}_2$ is dependent on all $z_1$, $z_2$ and $z_3$.

We formally describe the gate partition as following. Given a set of gates $G=\{g_1,g_2,\cdots,g_{|G|}\}$, let $B=\{b_1,b_2,\cdots,b_{|B|}\}$ be a partition of $G$, where $b_i$ denotes block $i$, and $|B|$ is the total number of blocks. Then we can approximate the variational posterior of $\bs{z}$ by modeling the distribution over blocks as
\[
q(\bs{z}) \approx q(b_1) q(b_2|b_1) q(b_3|b_1,b_2)\cdots q(b_{|B|}|b_{1},\cdots,b_{|B|-1}).
\]
To reduce the complexity, we can further simplify it as
\begin{equation}\label{eq:gate partition}
q(\bs{z}) \approx q(b_1) q(b_2|b_1) q(b_3|b_2)\cdots q(b_{|B|}| b_{|B|-1}),
\end{equation}
where block $i$ only depends on previous block $i-1$, ignoring all the other previous blocks, i.e., $q(b_i|b_{i-1})$, -- the first-order Markov assumption.


In our experiments, we define a layer-wise partition, i.e., a block containing all the filters in one layer. For example, in VGG16 after performing a layer-wise gate partition, we only need to model the dependency within 16 blocks instead of 4,224 gates, and therefore the computational overhead can be reduced significantly.

\subsection{Neural Dependency Modeling}
\label{sec:Neural Dependency Modeling}
Until now we have discussed the dependency modeling in a mathematical form. To incorporate the dependency modeling into the original deep network, we adopt neural networks to model the dependencies among gates. Specifically, we choose to use an MLP network as the \textit{gate generator}. With the proposed \textit{layer-wise} gate partition (i.e., attaching an MLP layer to a convolutional layer and a gate to a filter; see Fig.~\ref{fig: gate partition}(b)), the MLP architecture can model the dependency of gates, as expressed in Eq.~\ref{eq:gate partition}, effectively. 

Formally, we represent the gate generator as an MLP, with $gen_l$ denoting the operation of the $l$th layer. The binary gate $z_{lk}$ (i.e. the $k$th gate in block $l$) can be generated by 
\begin{align}\label{eq:mlp}
  &\log\alpha_{0}=\bs{1}\nonumber\\
  &\log\alpha_l = gen_l (\log\alpha_{l-1})\;\;\;\text{with } gen_l(\cdot) = c \cdot \tanh (W_l \cdot),\nonumber\\
  &z_{lk} \sim \text{HC}(\log\alpha_{lk},\beta),
\end{align}
where $W_l$ is the weight matrix of MLP at the $l$th layer, $c$ is a hyperparameter that bounds the value of $\log\alpha$ in the range of $(-c, c)$, and $\text{HC}(\log\alpha,\beta)$ is the Hard Concrete distribution with the location parameter $\log\alpha$ and the temperature parameter $\beta$~\cite{louizos2017learning}\footnote{Following $L_0$-HC~\cite{louizos2017learning}, $\beta$ is fixed to 2/3 in our experiments.}, which makes the sample $z_{lk}$ differentiable w.r.t. $\log\alpha_{lk}$. We set $c=10$ as default, which works well in all our experiments.

\begin{figure}[t]
  \begin{center}
    \includegraphics[width=0.9\linewidth]{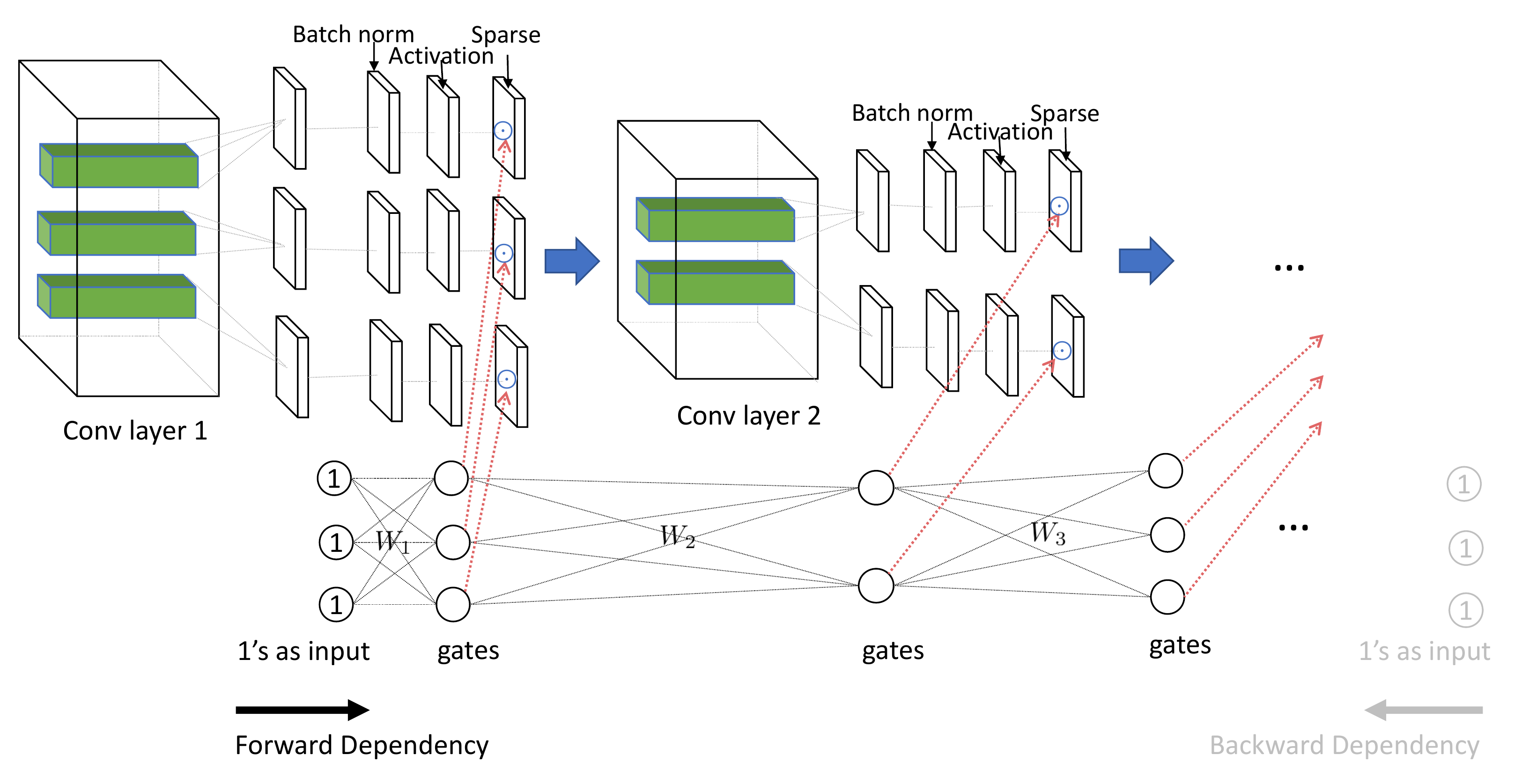}
  \end{center}\vspace{-15pt}
  \caption{The computational graph of Dep-$L_0$. The original CNN network is shown on the top, and the gate generator network (MLP) is shown at the bottom. Instead of attaching gates directly to filters, we attach gates to the feature maps after batch normalization (as shown by the red dotted lines). The gate can be generated by propagating the generator networks in either forward or backward direction. Both the original network and the gate generator are trained together as the whole pipeline is fully differentiable. $\{W_1,W_2, W_3,\cdots\}$ are the parameters of the MLP gate generator.}\label{fig:arch}
\end{figure}

Fig.~\ref{fig:arch} illustrates the overall architecture of Dep-$L_0$. The original network is shown on the top, and the gate generator (MLP) is shown at the bottom. Here we have a layer-wise partition of the gates, so the gate generator has the same number of layers as the original network. As we discussed in group sparsity, each gate is attached to a filter's output feature map after batch normalization. The input of the gate generator is initialized with a vector of 1's (i.e., $\log\alpha_0=\bs{1}$, such that all the input neurons are activated at the beginning). The values of gates $\bs{z}$ are generated as we forward propagate the generator. The generated $\bs{z}$s are then attached to original networks, and the gate dependencies can be learned from the data directly. The whole pipeline (the original network and the gate generator) is fully differentiable as the Hard Concrete distribution (instead of Bernoulli) is used to sample $\bs{z}$, so that we can use backpropagation to optimize the whole pipeline. 

Furthermore, as shown in Fig.~\ref{fig:arch}, the dependencies can be modeled in a backward direction as well, i.e., we generate the gates from the last layer $L$ of MLP first, and then generate the gates from layer $L-1$, and so on. In our experiments, we will evaluate the performance impacts of both forward and backward modeling.

In addition to MLPs, other network architectures such as LSTMs and CNNs can be used to model the gate generator as well. However, neither LSTMs nor CNNs achieves a competitive performance to MLPs in our experiments. Detailed ablation study is provided in the appendix.

\section{Experiments}

In this section we compare Dep-$L_0$ with the state-of-the-art pruning algorithms for CNN architecture pruning. In order to demonstrate the generality of Dep-$L_0$, we consider multiple image classification benchmarks (CIFAR10, CIFAR100~\cite{krizhevsky2009learning} and ImageNet~\cite{deng2009imagenet}) and multiple modern CNN architectures (VGG16~\cite{simonyan2014very}, ResNet50, and ResNet56~\cite{he2016deep}). As majority of the computations of modern CNNs are in the convolutional layers, following the competing pruning methods, we only prune the convolutional filters and leave the fully connected layers intact (even though our method can be used to prune any layers of a network). For a fair comparison, our experiments closely follow the benchmark settings provided in the literature. All our experiments are performed with PyTorch on Nvidia V100 GPUs. The details of the experiment settings can be found in the appendix.

\paragraph{\textbf{$L_0$-HC implementations}}
From our experiments, we found that the original $L_0$-HC implementation\footnote{\url{https://github.com/AMLab-Amsterdam/L0\_regularization}} has a couple issues. First, the binary gates are not properly attached after batch normalization, which results in pruned neurons still having impact after being removed. Second, it only uses one optimizer -- Adam for the original network parameters and the hard concrete parameters. We noted that using two optimizers: SGD with momentum for the original network and Adam for the hard concrete parameters works better. Therefore, we fixed these issues of $L_0$-HC for all the experiments and observed improved performance. For a fair comparison, we follow the same experiment settings as in Dep-$L_0$, and tune $L_0$-HC for the best performance.


\subsection{CIFAR10 Results}
We compare Dep-$L_0$ with ten state-of-the-art filter pruning algorithms, including our main baseline $L_0$-HC, in this experiment. Since the baseline accuracies in all the reference papers are different, we compare the performances of all competing methods by their accuracy gains $\Delta_{Acc}$ and their pruning rates in terms of FLOPs and network parameters. For Dep-$L_0$, we evaluate our algorithm with \textit{forward} and \textit{backward} dependency modeling. Table~\ref{tab:cifar10} provides the results on CIFAR10. As can be seen, for VGG16, our algorithm (with \textit{backward} dependency modeling) achieves the highest FLOPs reduction of 65.9\% on CIFAR10 with only 0.1\% of accuracy loss. For ResNet56, our \textit{forward} dependency modeling achieves the highest accuracy gain of 0.2\% with a very competitive FLOPs reduction of 45.5\%. 

\begin{table}[ht]
  \small
  \centering
  \caption{Comparison of pruning methods on CIFAR10. ``$\Delta_{Acc}$": `+' denotes accuracy gain; `-' denotes accuracy loss; the worst result is in red. ``FLOPs (P.R. \%)": pruning ratio in FLOPs. ``Params. (P.R. \%)": prune ratio in parameters. ``-": results not
   reported in original paper.}
  \begin{tabular}{l l c c c c}\toprule
      Model & Method & Acc. (\%) & $\Delta_{Acc}$ & FLOPs (P.R. \%) & Params. (P.R. \%) \\\hline
      \multirow{7}{*}{VGG16} & Slimming~\cite{liu2017learning} & 93.7$\rightarrow$93.8 &+0.1 &195M (51.0) & 2.30M (88.5) \\
       & DCP~\cite{zhuang2018discrimination} & 94.0$\rightarrow$94.6 & \textbf{+0.6} & 109.8M (65.0) & \textbf{0.94M (93.6)}\\
       & AOFP~\cite{ding2019approximated} & 93.4$\rightarrow$93.8 & +0.4 &215M (31.3) & -  \\
       & HRank~\cite{lin2020hrank} & 94.0$\rightarrow$93.4 &\textcolor{red}{-0.6} & 145M (53.5) & 2.51M (82.9)\\
       & \cellcolor{blue!25}$L_0$-HC (Our implementation) & \cellcolor{blue!25}93.5$\rightarrow$93.1  & \cellcolor{blue!25}-0.4 & \cellcolor{blue!25}135.6M (39.8) & \cellcolor{blue!25}2.8M (80.9) \\
       & \cellcolor{blue!25}Dep-$L_0$ (forward) & \cellcolor{blue!25}93.5$\rightarrow$93.5 & \cellcolor{blue!25}0 & \cellcolor{blue!25}111.9M (64.4) & \cellcolor{blue!25}2.1M  (85.7) \\
       & \cellcolor{blue!25}Dep-$L_0$ (backward) & \cellcolor{blue!25}93.5$\rightarrow$93.4 & \cellcolor{blue!25}-0.1 & \cellcolor{blue!25}\textbf{107.0M (65.9)} & \cellcolor{blue!25}1.8M (87.8) \\
       \hline
      \multirow{9}{*}{ResNet56} & SFP~\cite{he2018soft} & 93.6$\rightarrow$93.4 & -0.2 & 59.4M (53.1) & - \\
       & AMC~\cite{he2018amc} & 92.8$\rightarrow$91.9 & -0.9 & 62.5M (50.0) & - \\
       & DCP~\cite{zhuang2018discrimination} & 93.8$\rightarrow$93.8 & 0 &67.1M (47.1) & \textbf{0.25M (70.3)}\\ 
       & FPGM~\cite{he2019filter} & 93.6$\rightarrow$93.5 & -0.1 & 59.4M (52.6) & - \\
       & TAS~\cite{dong2019network} & 94.5$\rightarrow$93.7 &\textcolor{red}{-0.8} & \textbf{59.5M (52.7)} & - \\
       & HRank~\cite{lin2020hrank} & 93.3$\rightarrow$93.5 &\textbf{+0.2} & 88.7M (29.3) & 0.71M (16.8) \\     
       \cline{2-6}
       & \cellcolor{blue!25}$L_0$-HC (Our implementation)  & \cellcolor{blue!25}93.3$\rightarrow$92.8 &\cellcolor{blue!25}-0.5& \cellcolor{blue!25}71.0M (44.1) & \cellcolor{blue!25}0.46M (45.9)     \\
       & \cellcolor{blue!25}Dep-$L_0$ (forward) & \cellcolor{blue!25}93.3$\rightarrow$93.5 & \cellcolor{blue!25}\textbf{+0.2} & \cellcolor{blue!25}69.1M (45.5) & \cellcolor{blue!25}0.48M (43.5)     \\
       & \cellcolor{blue!25}Dep-$L_0$ (backward) & \cellcolor{blue!25}93.3$\rightarrow$93.0 & \cellcolor{blue!25}-0.3 & \cellcolor{blue!25}66.7M (47.4) & \cellcolor{blue!25}0.49M (42.4)     \\
       \bottomrule
  \end{tabular}
  \label{tab:cifar10}
\end{table}

\begin{table*}[ht!]
  \small
  \centering
  \caption{Comparison of pruning methods on CIFAR100. ``$\Delta_{Acc}$": `+' denotes accuracy gain; `-' denotes accuracy loss; the worst result is in red. ``FLOPs (P.R. \%)": pruning ratio in FLOPs. ``Params. (P.R. \%)": prune ratio in parameters. ``-": results not
  reported in original paper.}
  \begin{tabular}{l l c c c c}\toprule
     Model & Method & Acc. (P.R. \%) & $\Delta_{Acc}$  & FLOPs (P.R. \%) & Params. (\%) \\\hline
      \multirow{4}{*}{VGG16} & Slimming~\cite{liu2017learning} & 73.3$\rightarrow$73.5 & +0.2& 250M (37.1) & 5.0M (75.1) \\
      & \cellcolor{blue!25}$L_0$-HC (Our implementation) & \cellcolor{blue!25}72.2$\rightarrow$70.0 & \cellcolor{blue!25}\textcolor{red}{-1.2} &  \cellcolor{blue!25}138M (56.2) & \cellcolor{blue!25}4.1M (72.5) \\
      & \cellcolor{blue!25}Dep-$L_0$ (forward) & \cellcolor{blue!25}72.2$\rightarrow$71.6 & \cellcolor{blue!25}-0.6 & \cellcolor{blue!25}\textbf{98M (68.8)} & \cellcolor{blue!25}\textbf{2.1M  (85.7) }\\
      & \cellcolor{blue!25}Dep-$L_0$ (backward) & \cellcolor{blue!25}72.2$\rightarrow$72.5 & \cellcolor{blue!25}\textbf{+0.3} &\cellcolor{blue!25}105M (66.6) & \cellcolor{blue!25}2.2M (85.0) \\\hline
      \multirow{6}{*}{ResNet56} & SFP~\cite{he2018soft} & 71.4$\rightarrow$68.8 & \textcolor{red}{-2.6} & \textbf{59.4M (52.6)} & -\\
      & FPGM~\cite{he2019filter} & 71.4$\rightarrow$69.7 & -1.7 & \textbf{59.4M (52.6)} & -\\
      & TAS~\cite{dong2019network} & 73.2$\rightarrow$72.3 & -0.9 & 61.2M (51.3) & -\\
      & \cellcolor{blue!25}$L_0$-HC (Our implementation) & \cellcolor{blue!25}71.8$\rightarrow$70.4 & \cellcolor{blue!25}-1.4  & \cellcolor{blue!25}82.2M (35.2) & \cellcolor{blue!25}0.73M (15.2) \\
      & \cellcolor{blue!25}Dep-$L_0$ (forward) & \cellcolor{blue!25}71.8$\rightarrow$71.7 & \cellcolor{blue!25}\textbf{-0.1} & \cellcolor{blue!25}87.6M (30.9) & \cellcolor{blue!25}0.56M (34.9) \\
      & \cellcolor{blue!25}Dep-$L_0$ (backward) & \cellcolor{blue!25}71.8$\rightarrow$71.2 & \cellcolor{blue!25}-0.6 & \cellcolor{blue!25}93.4M (26.3) & \cellcolor{blue!25}\textbf{0.52M (39.5)} \\
      \bottomrule
  \end{tabular}
  \label{tab:cifar100}
\end{table*}

Since $L_0$-HC is our main baseline, we highlight the comparison between Dep-$L_0$ and $L_0$-HC in the table. As we can see, Dep-$L_0$ outperforms $L_0$-HC consistently in all the experiments. For VGG16, $L_0$-HC prunes only 39.8\% of FLOPs but suffers from a 0.4\% of accuracy drop, while our algorithm prunes more (65.9\%) and almost keeps the same accuracy (-0.1\%). For ResNet56, our algorithm prunes more (45.5\% v.s. 44.1\%) while achieves a higher accuracy (0.2\% vs. -0.5\%) than that of $L_0$-HC. 

\subsection{CIFAR100 Results}
Experimental results on CIFAR100 are reported in Table~\ref{tab:cifar100}, where Dep-$L_0$ is compared with four state-of-the-arts pruning algorithms: Slimming~\cite{liu2017learning}, SFP~\cite{he2018soft}, FPGM~\cite{he2019filter} and TAS~\cite{dong2019network}. Similar to the results on CIFAR10, on this benchmark Dep-$L_0$ achieves the best accuracy gains and very competitive or sometimes even higher prune rates compared to the state-of-the-arts. More importantly, Dep-$L_0$ outperforms $L_0$-HC in terms of classification accuracies and pruning rates consistently, demonstrating the effectiveness of dependency modeling.

\subsection{ImageNet Results}
The main goal of the paper is to make $L_0$-HC once again competitive on the large-scale benchmark of ImageNet. In this section, we conduct a comprehensive experiment on ImageNet, where the original $L_0$-HC fails to prune without a significant damage of model quality~\cite{gale2019state}. Table~\ref{tab:imagenet} reports the results on ImageNet, where eight state-of-the-art filter pruning methods are included, such as SSS-32~\cite{huang2018data}, DCP~\cite{zhuang2018discrimination}, Taylor~\cite{molchanov2019importance}, FPGM~\cite{he2019filter}, HRank~\cite{lin2020hrank} and others. As can be seen, Dep-$L_0$ (forward) prunes 36.9\% of FLOPs and 37.2\% of parameters with a 1.38\% of accuracy loss, which is comparable with other state-of-the-art algorithms as shown in the table. 
\begin{table}[t]
  \small
  \centering
  \caption{Comparison of pruning methods on ImageNet. ``$\Delta_{Acc}$": `+' denotes accuracy gain; `-' denotes accuracy loss; the worst result is in red. ``FLOPs (P.R. \%)": pruning ratio in FLOPs. ``Params. (P.R. \%)": prune ratio in parameters. ``-": results not
   reported in original paper.}

  \begin{tabular}{l l c c c c}\toprule
    Model & Method &  Acc. (\%) & $\Delta_{Acc}$ & FLOPs (P.R.\%) & Params. (P.R.\%) \\\hline
      \multirow{13}{*}{ResNet50}  & SSS-32~\cite{huang2018data} &  76.12 $\rightarrow$ 74.18 & -1.94 & 2.82B (31.1) & 18.60M (27.3) \\
      & DCP~\cite{zhuang2018discrimination} & 76.01 $\rightarrow$ 74.95 & -1.06 & \textbf{1.82B (55.6)} & \textbf{12.40M (51.5)}\\
      & GAL-0.5~\cite{lin2019towards} & 76.15 $\rightarrow$ 71.95 & \textcolor{red}{-4.2} & 2.33B (43.1) & 21.20M (17.2) \\
      & Taylor-72~\cite{molchanov2019importance} & 76.18 $\rightarrow$ 74.50 & -1.68 & 2.25B (45.0) & 14.20M (44.5) \\
      & Taylor-81~\cite{molchanov2019importance} & 76.18 $\rightarrow$ 75.48 & -0.70 & 2.66B (34.9) & 17.90M (30.1) \\
      & FPGM-30~\cite{he2019filter} & 76.15 $\rightarrow$ 75.59 & -0.56 & 2.36B (42.2) & - \\
      & FPGM-40~\cite{he2019filter} & 76.15 $\rightarrow$ 74.83 & -1.32 & 1.90B (53.5) & - \\
      & LeGR~\cite{chin2019legr} & 76.10 $\rightarrow$ 75.70 & -0.40 & 2.37B (42.0) & - \\ 
      & TAS~\cite{dong2019network} & 77.46 $\rightarrow$ 76.20 & -1.26 & 2.31B (43.5) & - \\
      & HRank~\cite{lin2020hrank} & 76.15 $\rightarrow$ 74.98 & -1.17 & 2.30B (43.8) & 16.15M (36.9) \\
      & \cellcolor{blue!25}$L_0$-HC (Our implementation) & \cellcolor{blue!25}76.15 $\rightarrow$ 76.15 & \cellcolor{blue!25}0 & \cellcolor{blue!25}4.09B (\textcolor{red}{0.00}) & \cellcolor{blue!25}25.58M (\textcolor{red}{0.00}) \\
      & \cellcolor{blue!25}Dep-$L_0$ (forward) & \cellcolor{blue!25}76.15  $\rightarrow$ 74.77 & \cellcolor{blue!25}-1.38 & \cellcolor{blue!25}2.58B (36.9) & \cellcolor{blue!25}16.04M (37.2) \\
      & \cellcolor{blue!25}Dep-$L_0$ (backward) & \cellcolor{blue!25}76.15  $\rightarrow$ 74.70 & \cellcolor{blue!25}-1.45 & \cellcolor{blue!25}2.53B (38.1) & \cellcolor{blue!25}14.34M (43.9) \\
      \bottomrule
  \end{tabular}
  \label{tab:imagenet}
\end{table}

Again, since $L_0$-HC is our main baseline, we highlight the comparison between Dep-$L_0$ and $L_0$-HC in the table. We tune the performance of $L_0$-HC extensively by searching for the best hyperparameters in a large  space. However, even with extensive efforts, $L_0$-HC still fails to prune the network without a significant damage of model quality, confirming the observation made by~\cite{gale2019state}. On the other hand, our Dep-$L_0$ successfully prunes ResNet50 with a very competitive pruning rate and high accuracy compared to the state-of-the-arts, indicating that our dependency modeling indeed makes the original $L_0$-HC very competitive on the large-scale benchmark of ImageNet -- the main goal of the paper.

\begin{figure}[ht]
  \begin{center}
    \subfigure[VGG16-C10 (Dep-$L_0$)]{\includegraphics[width=0.32\linewidth]{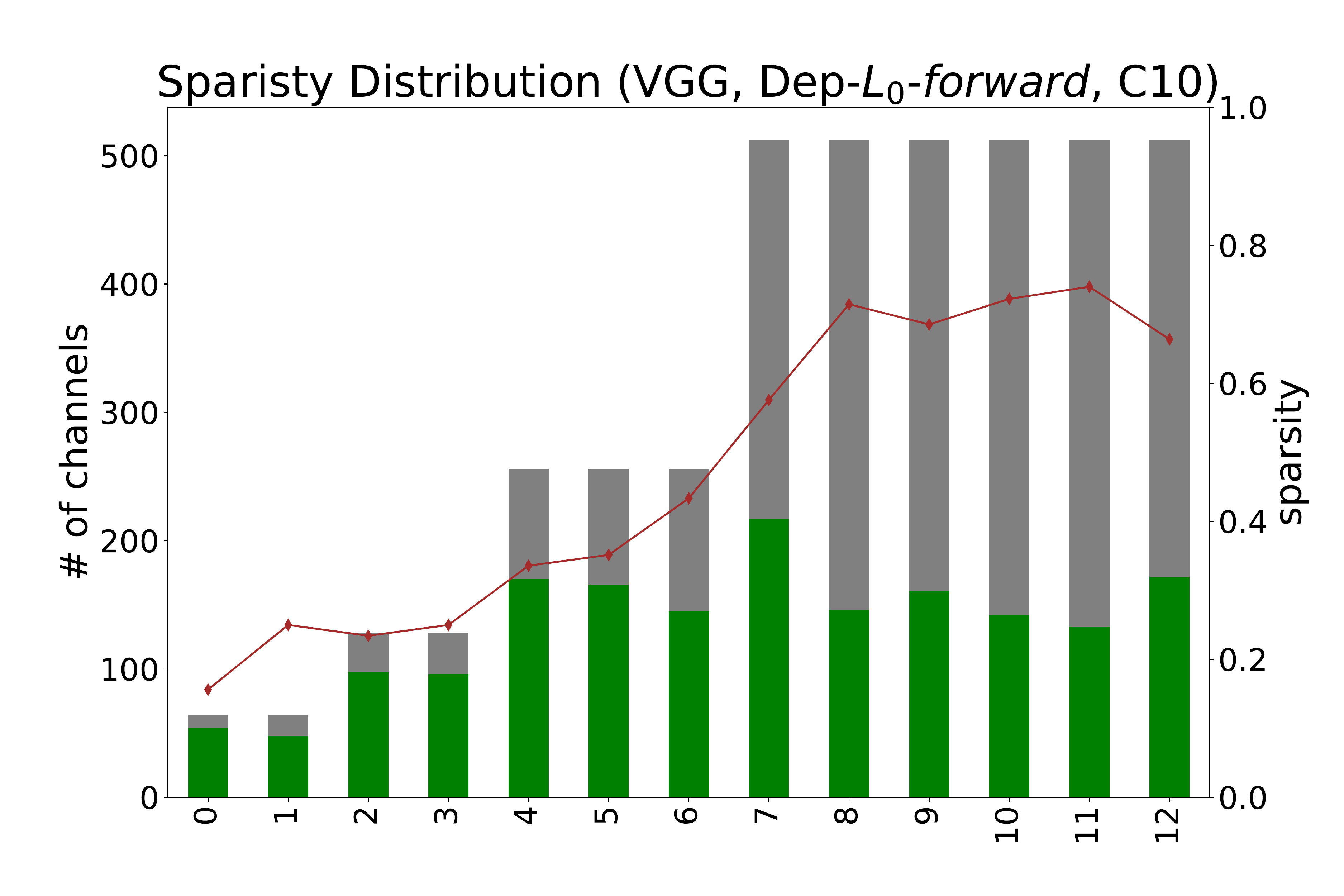}}
    \subfigure[VGG16-C10 ($L_0$-HC)]{\includegraphics[width=0.32\linewidth]{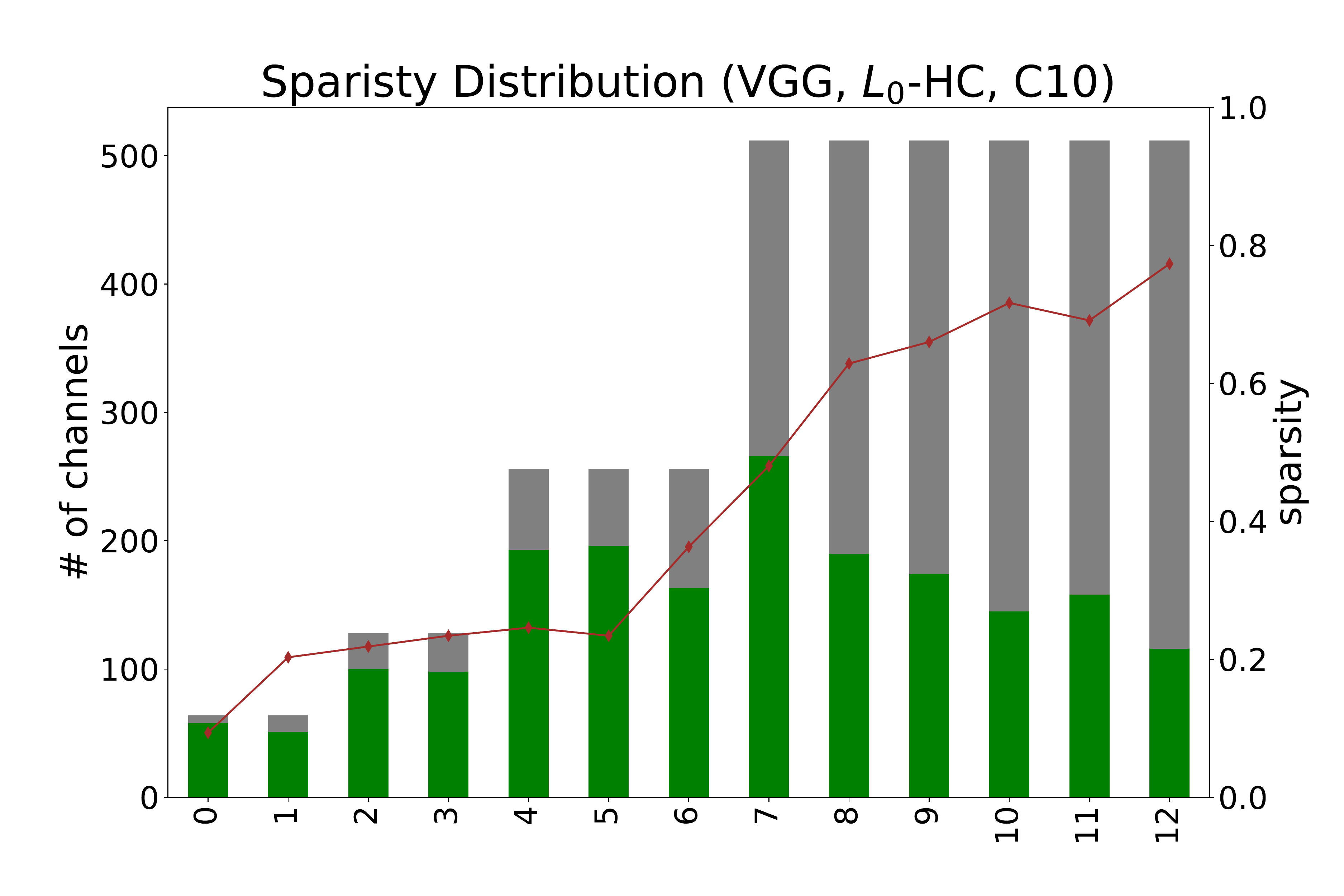}}
    \subfigure[R56-C100 (Dep-$L_0$)]{\includegraphics[width=0.32\linewidth]{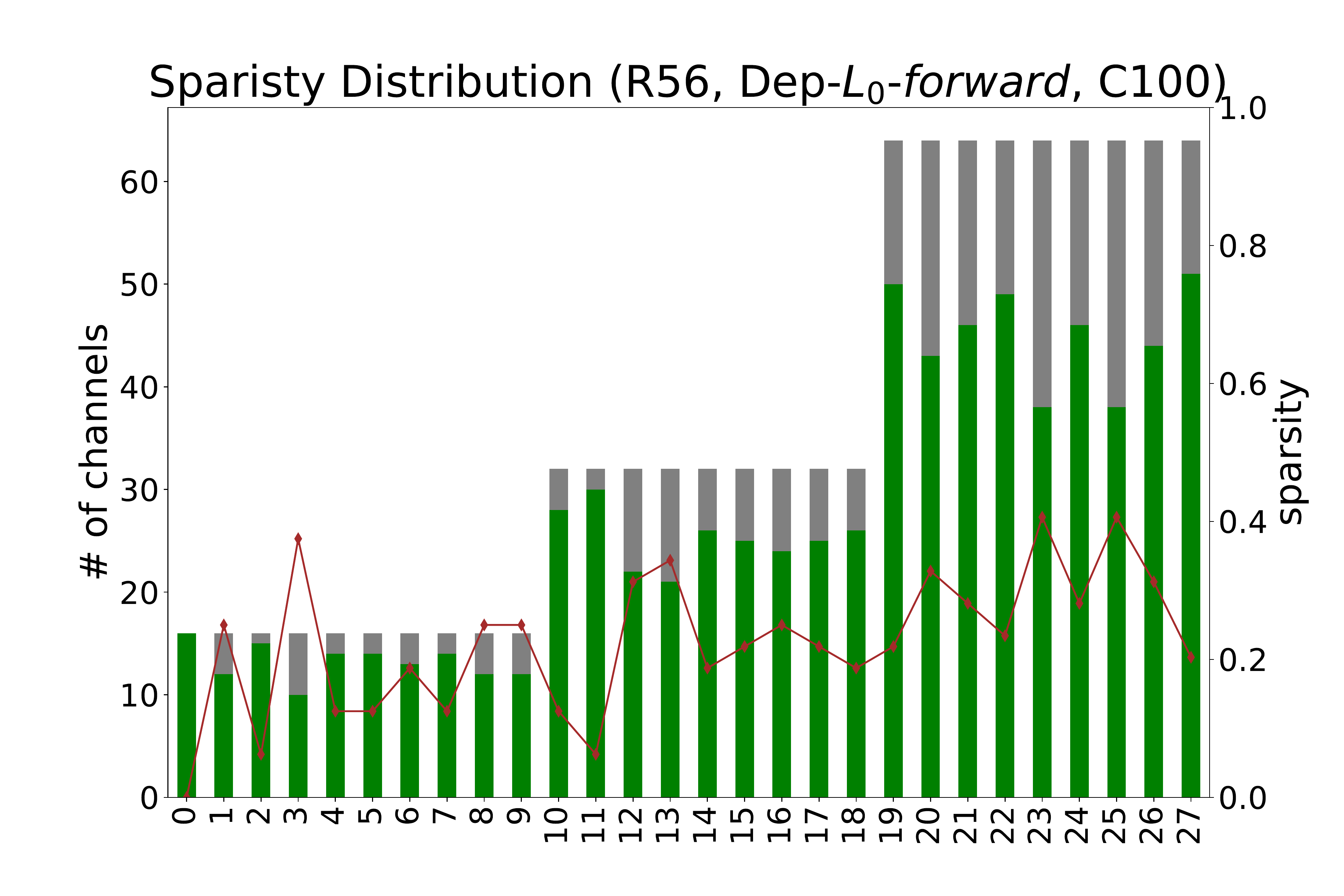}}\hfill
    \subfigure[R56-C100 ($L_0$-HC)]{\includegraphics[width=0.32\linewidth]{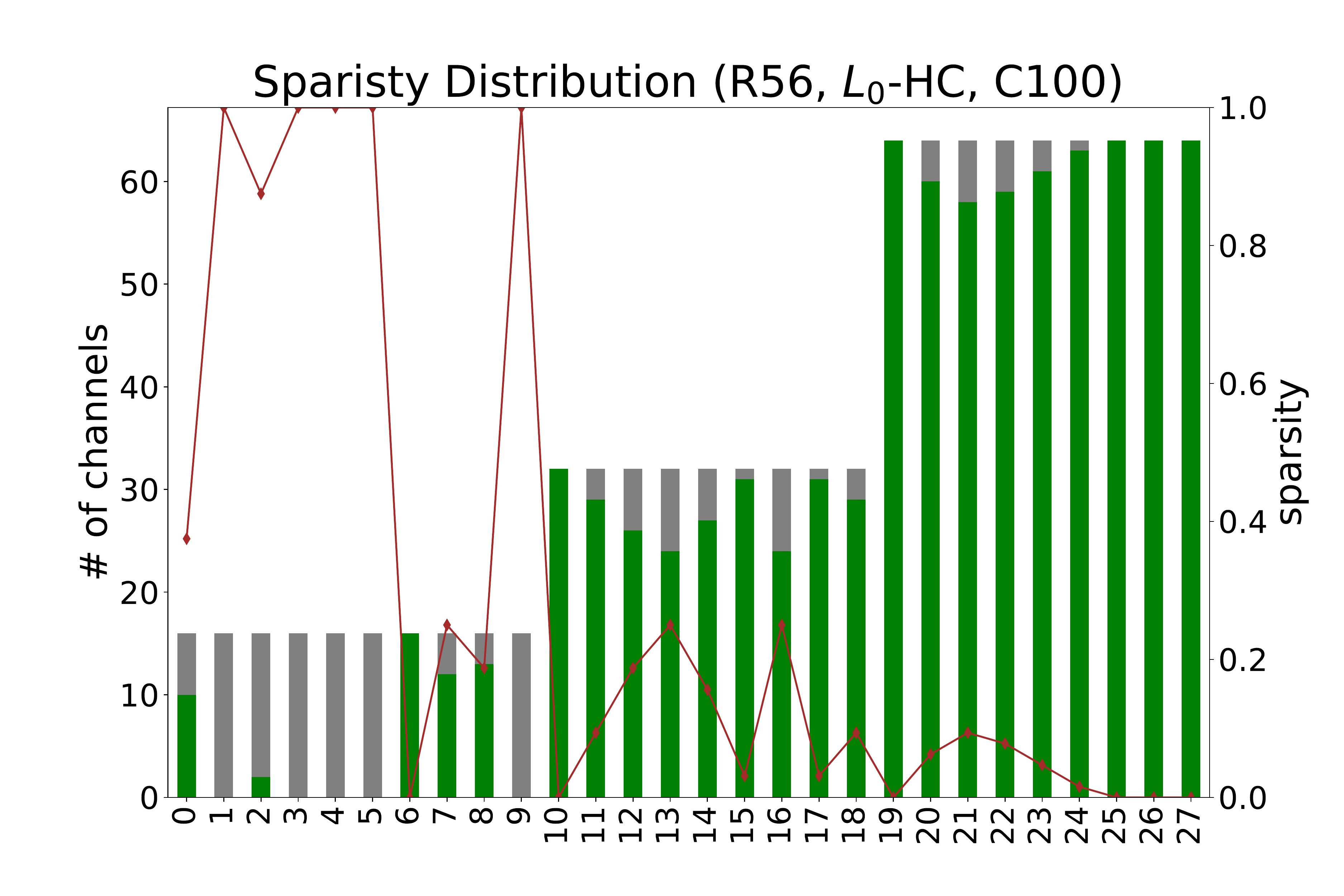}} 
    \subfigure[R50-ImageNet (Dep-$L_0$)]{\includegraphics[width=0.32\linewidth]{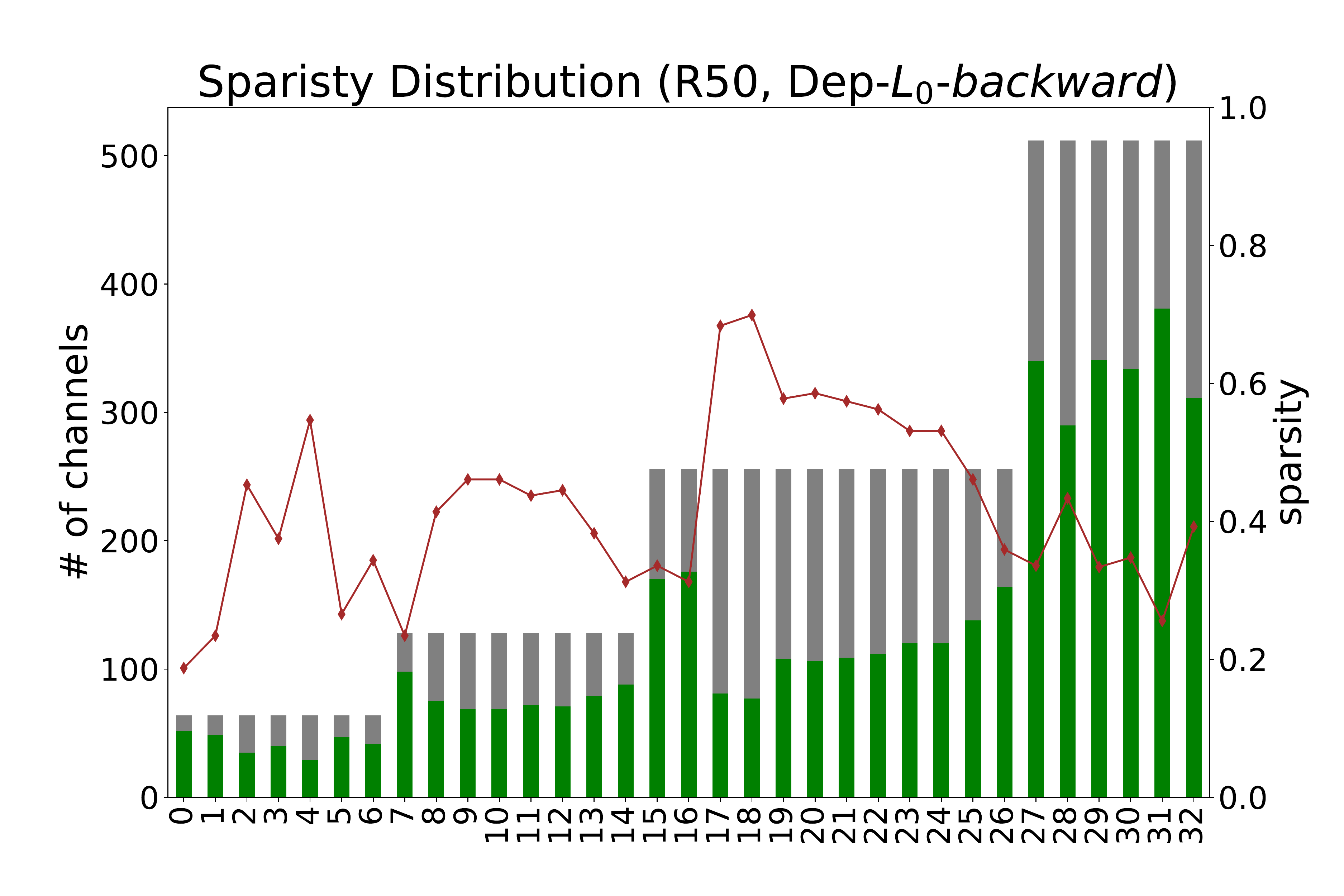}}
    \subfigure[R50-ImageNet ($L_0$-HC)]{\includegraphics[width=0.32\linewidth]{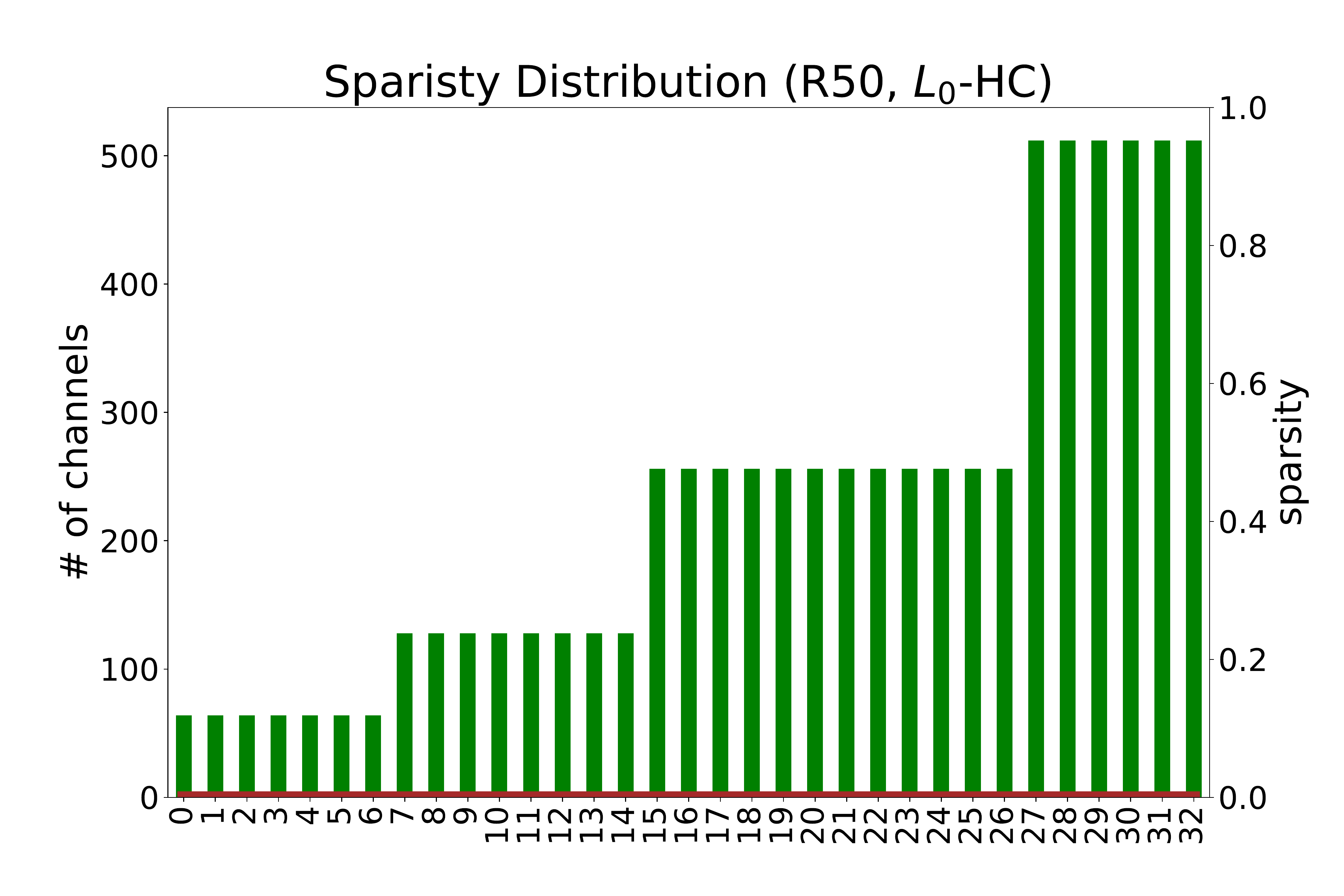}}
  \end{center}\vspace{-10pt}
  \caption{The layer-wise prune ratios (red curves) of learned sparse structures. The height of a bar denotes  the number of filters of a convolutional layer and gray (green) bars correspond to the original (pruned) architecture, respectively. ``R50/R56": ResNet 50/56; ``C10/C100": CIFAR10/100.}\label{fig: dist of sparsity}
\end{figure}

\subsection{Study of Learned Sparse Structures}
To understand of the behavior of Dep-$L_0$, we further investigate the sparse structures learned by Dep-$L_0$ and $L_0$-HC, with the results reported in Fig.~\ref{fig: dist of sparsity}. For VGG16 on CIFAR10, Figs.~\ref{fig: dist of sparsity}(a-b) demonstrate that both algorithms learn a similar sparsity pattern: the deeper a layer is, the higher prune ratio is, indicating that the shallow layers of VGG16 are more important for its predictive performance. However, for deeper networks such as ResNet56 and ResNet50, the two algorithms perform very differently. For ResNet56 on CIFAR100, Figs.~\ref{fig: dist of sparsity}(c-d) show that Dep-$L_0$ sparsifies each layer by a roughly similar prune rate: it prunes around 20\% of the filters in first 12 layers, and around 30\% of the filters in the rest of layers. However, on the same benchmark $L_0$-HC tends to prune \textit{all or nothing}: it completely prunes 5 out of 28 layers, but does not prune any filters in other six layers; for the rest of layers, the sparsity produced by $L_0$-HC is either extremely high or low. As of ResNet50 on ImageNet, Figs.~\ref{fig: dist of sparsity}(e-f) show that the difference between Dep-$L_0$ and $L_0$-HC is more significant: Dep-$L_0$ successfully prunes the model with a roughly similar prune rate across all convolutional layers, while $L_0$-HC fails to prune any filters. 

\subsection{Run-time Comparison}

\begin{table}[ht]
  \small
  \centering
  \caption{Run-time comparison between Dep-$L_0$ and $L_0$-HC. ``R50/R56": ResNet50/56; ``C10/C100": CIFAR10/100. ``BC": Before Convergence; ``TTS": Time to Solution.}

  \begin{tabular}{l l c c c c}\toprule
      Benchmark & Method & \# Epochs & \# Epochs BC & Per-epoch Time BC & TTS\\ 
      \hline
      & $L_0$-HC & 300 & 218 & 37.9 sec & 160 min\\ 
      R56-C10 & Dep-$L_0$ (forward) & 300 & \textbf{106} & 42.2 sec & \textbf{124 min} \\
      &  Dep-$L_0$ (backward) & 300 & 140 &42.6 sec & 141 min \\
      \hline
      & $L_0$-HC & 300 & 117 & 38.7 sec & 167 min \\
      R56-C100 & Dep-$L_0$ (forward) & 300 & \textbf{58} & 43.9 sec & \textbf{127 min}  \\
      & Dep-$L_0$ (backward) & 300 & 61 &43.6 sec & 133 min \\
      \hline
      & $L_0$-HC & 90 & fail to prune & 4185 sec & 104.6 h\\
      R50-ImageNet & Dep-$L_0$ (forward) & 90 & \textbf{30} &4342 sec & \textbf{59.5 h} \\
      & Dep-$L_0$ (backward) & 90 & 32 & 4350 sec &  60.1 h\\
      \bottomrule
  \end{tabular}
  \label{tab: converge}
\end{table}

The main architectural difference between Dep-$L_0$ and $L_0$-HC is the gate generator. Even though the gate generator (MLP) is relatively small compared to the original deep network to be pruned, its existence increases the computational complexity of Dep-$L_0$. Thus, it is worth comparing the run-times of Dep-$L_0$ and $L_0$-HC as well as their convergence rates in terms of sparse structure search. Once a sparse structure is learned by a pruning algorithm, we can extract the sparse network from the original network and continue the training on the smaller structure such that we can reduce the total time to solution (TTS). To this end, we compare Dep-$L_0$ with $L_0$-HC in terms of (1) structure search convergence rate, i.e., how many training epochs are needed for a pruning algorithm to converge to a sparse structure? (2) Per-epoch training time before convergence, and (3) the total time to solution (TTS). The results are reported in Table~\ref{tab: converge}. As can be seen, Dep-$L_0$ (both \textit{forward} and \textit{backward}) converges to a sparse structure in roughly half of the epochs that $L_0$-HC needs (column 4). Even though the per-epoch training time of Dep-$L_0$ is 12\% (4\%) larger than that of $L_0$-HC on CIFAR10/100 (ImageNet) due to the extra computation of the gate generator (column 5), the total time to solution reduces by 22.5\% (43.1\%) on the CIFAR10/100 (ImageNet) benchmarks thanks to the faster convergence rates and sparser models induced by Dep-$L_0$ as compared to $L_0$-HC (column 6).

\section{Conclusion and Future Work}

We propose Dep-$L_0$, an improved $L_0$ regularized network sparsification algorithm via dependency modeling. The algorithm is inspired by a recent observation of Gale et al.~\cite{gale2019state} that $L_0$-HC performs inconsistently in large-scale learning tasks. Through the lens of variational inference, we found that this is likely due to the mean-field assumption in variational inference that ignores the dependency among all the neurons for network pruning. We further propose a dependency modeling of binary gates to alleviate the deficiency of the original $L_0$-HC. A series of experiments are performed to evaluate the generality of our Dep-$L_0$. The results show that our Dep-$L_0$ outperforms the original $L_0$-HC in all the experiments consistently, and the dependency modeling makes the $L_0$-based sparsification once again very competitive and sometimes even outperforms the state-of-the-art pruning algorithms. Further analysis shows that Dep-$L_0$ also learns a better structure in fewer epochs, and reduces the total time to solution by 20\%-40\%. 

As for future work, we plan to explore whether dependency modeling can be used to improve other pruning methods. To the best of our knowledge, there are very few prior works considering dependency for network pruning (e.g.,~\cite{park2020lookahead}). Our results show that this may be a promising direction to further improve many existing pruning algorithms. Moreover, the way we implement dependency modeling is still very preliminary, which can be improved further in the future. 

\vspace{-4pt}
\section*{Acknowledgment} 
We would like to thank the anonymous reviewers for their comments and suggestions, which helped improve the quality of this paper. We would also gratefully acknowledge the support of VMware Inc. for its university research fund to this research.

\appendix

\section{Experimental Details} 
The experimental details are provided in this section for the purpose of reproducibility. For a fair comparison, our experiments closely follow the benchmark settings provided in the literature. 

\paragraph{VGG16 on CIFAR10/100}
We adopt a tailored VGG16 network~\cite{liu2015very} to the CIFAR10/100 datasets. The network contains 13 convolutional layers and 1 fully connected layer. To sparisfy the convolutional layers, the binary gates are attached to the output feature maps of all convolutional filters and the dependencies of the gates are modeled by an MLP in either \textit{forward} or \textit{backward} direction, as described in Sec.~3.4. The models are trained on CIFAR10/100 for 300 epochs with a batch size of 128. We apply two different optimizers for two different groups of parameters: SGD with momentum for VGG16 parameters, and Adam~\cite{kingma2014adam} for the parameters of the gate generator. The initial learning rate is set to 0.05 for SGD and 0.001 for Adam. The momentum and weight decay is set to 0.9 and 5e-4, respectively. The learning rate is multiplied by 0.2 every 60 epochs. For the gate generator, we initialize the bias terms of each MLP layer with samples from a Gaussian distribution $\mathcal{N}(3, 0.01)$ to encourage gates being activated at the beginning of training.


\paragraph{ResNet56 on CIFAR10/100}
To evaluate the effectiveness of our Dep-$L_0$ on a more compact and modern CNN architecture, we further conduct experiments on ResNet56~\cite{he2016deep}. Due to the existence of skip connections, we do not sparsify the last convolutional layer of each residual block to keep a valid addition operation. The dependencies of gates across layers are then built as an MLP. The training has been done in 300 epochs. Two optimizers are used: SGD with momentum for ResNet56 parameters with an initial learning rate of 0.1 and Adam for the gate generator parameters with an initial learning rate of 0.001. The batch size, weight decay and momentum of SGD are set to 128, 5e-4 and 0.9, respectively. The learning rate is multiplied by 0.1 and 0.2 every 60 epochs for SGD and Adam optimizer, respectively. 

\paragraph{ResNet50 on ImageNet}
Following~\cite{gale2019state}, we perform experiments on ImageNet~\cite{deng2009imagenet} with ResNet50 to evaluate our algorithm on a large-scale benchmark. This is one of our main experiments since Gale et al.~\cite{gale2019state} claim that $L_0$-HC fails to sparsify ResNet50 on ImageNet. Similar to the settings of ResNet56, we do not prune the last layer of each residual block. We train the model for 90 epochs and fine-tune for another 10 epochs with the gate generator frozen. We again use two optimizers: SGD with momentum as the optimizer for the parameters of ResNet50 with initial learning rate of 0.1 and Adam for the gate generator with initial learning rate of 0.001. The batch size, weight decay and momentum are set to 256, 1e-4 and 0.9, respectively. The learning rate is multiplied by 0.1 and 0.2 every 30 epochs for SGD and Adam optimizer, respectively. 

\section{Ablation Study of the Gate Generator}
In the main text, we model the dependency among the binary gates with an MLP network. We note that other network architectures, such as MLP variants, CNN and LSTM~\cite{gers1999learning}, can also be used to model the dependency. We therefore provide an ablation study to analyze their performances on VGG16-CIFAR10, with the results reported in Table~\ref{tab:ablation}.

\begin{table*}[ht!]
  \small
  \centering
  \caption{Ablation study of the gate generator architecture with VGG16 on CIFAR10. ``FLOPs": pruning ratio in FLOPs. ``Params.": prune ratio in parameters. }
  \begin{tabular}{l l c c c c}\toprule
      Dependency Modeling & Acc. (\%) & FLOPs (P.R. \%)  & Params. (P.R. \%)  \\\hline
       FC($c_1$, $c_2$) $\to$ selected model & 93.5 & 111.9M (64.4) & 2.1M (85.7) \\
       FC($c_1$, $c_2 * 2$) ReLU FC($c_2 * 2$, $c_2$) & 92.8 & 97.1M (69.1) & 2.4M (83.7) \\
       FC($c_1$, $c_2 * 2$) Tanh FC($c_2 * 2$, $c_2$) & 92.9 & 85.3M (72.8) &  1.3M (91.1) \\
       Conv1d ReLU FC & 93.4 & 162.3M (48.3) & 5.6M (61.9) \\
       Conv1d ReLU Conv1d ReLU Conv1d ReLU FC & 93.4 & 155.8M (50.4) & 5.4M (63.3) \\
       LSTM & not converged\\
       \bottomrule
  \end{tabular}
  \label{tab:ablation}
\end{table*}

\paragraph{MLP variants}
In Sec.~3.4, we propose to use a fully connected (FC) layer, parameterized by $W_l$, to connect the gates between two consecutive layers. Other variants of MLP can also be used to model the dependency. Suppose the two consecutive layers have $c_1$ and $c_2$ gates, respectively. The single-FC-layer model used in the main text can be denoted as $\text{FC}(c_1,c_2)$ (the 1st row of Table~\ref{tab:ablation}). Alternatively, we can also use two FC layers interspersed by ReLU or Tanh to represent the dependency among gates (the 2nd and 3rd rows of Table~\ref{tab:ablation}). As can be seen, these two MLP variants suffer from non-trivial accuracy drops, even though they achieve higher prune rates.

\paragraph{CNN}
We can also use a 1-D ConvNet as the gate generator. Specifically, we can model the dependency among gates in two consecutive layers by a 1-D convolution with the kernel size of $3\times 1$, followed by ReLU and a FC layer (the 4th row in Table~\ref{tab:ablation}). Moreover, deep 1-D ConvNet, such as the one listed in the 5th row of Table~\ref{tab:ablation} can be used. Even though they achieve similar accuracies as that of $\text{FC}(c_1,c_2)$, their prune rates are not competitive.

\paragraph{LSTM}
Finally, we exploit an LSTM as the gate generator to model the element-wise dependency autoregressively. We first experiment this LSTM gate generator with a small-scale CNN architecture -- LeNet5~\cite{lecun1998gradient}, which only contains 70 filters. We achieve a competitive pruning rate with almost no accuracy loss. However, when it comes to VGG16, which contains 4,224 filters, the LSTM gate generator has to predict 4,224 binary gates autoregressively. After exhaustive hyperparameter tuning, we still cannot get the LSTM gate generator converged and thus fail to sparsify VGG16. In addition, training of the LSTM is very time consuming due to the autoregressive modeling of 4,224 binary gates.

\paragraph{Summary}
The ablation study presented in Table~\ref{tab:ablation} show that $\text{FC}(c_1,c_2)$, the MLP network used in the main text, achieves the best balance between classification accuracy and the prune rate. Moreover, this MLP network is also computational efficient as it is relatively small compared to the original networks to be pruned. Therefore, we select $\text{FC}(c_1,c_2)$ as our MLP network in all the experiments.

\section{Sparse Structures Learned on CIFAR10}

\begin{figure}[h]
  \caption{The layer-wise prune rates after training ResNet56 on CIFAR10 with Dep-$L_0$ and $L_0$-HC.}\label{fig: dist of sparsity2}
    \begin{center}
      \subfigure[R56-C10 (Dep-$L_0$)]{\includegraphics[width=0.48\linewidth]{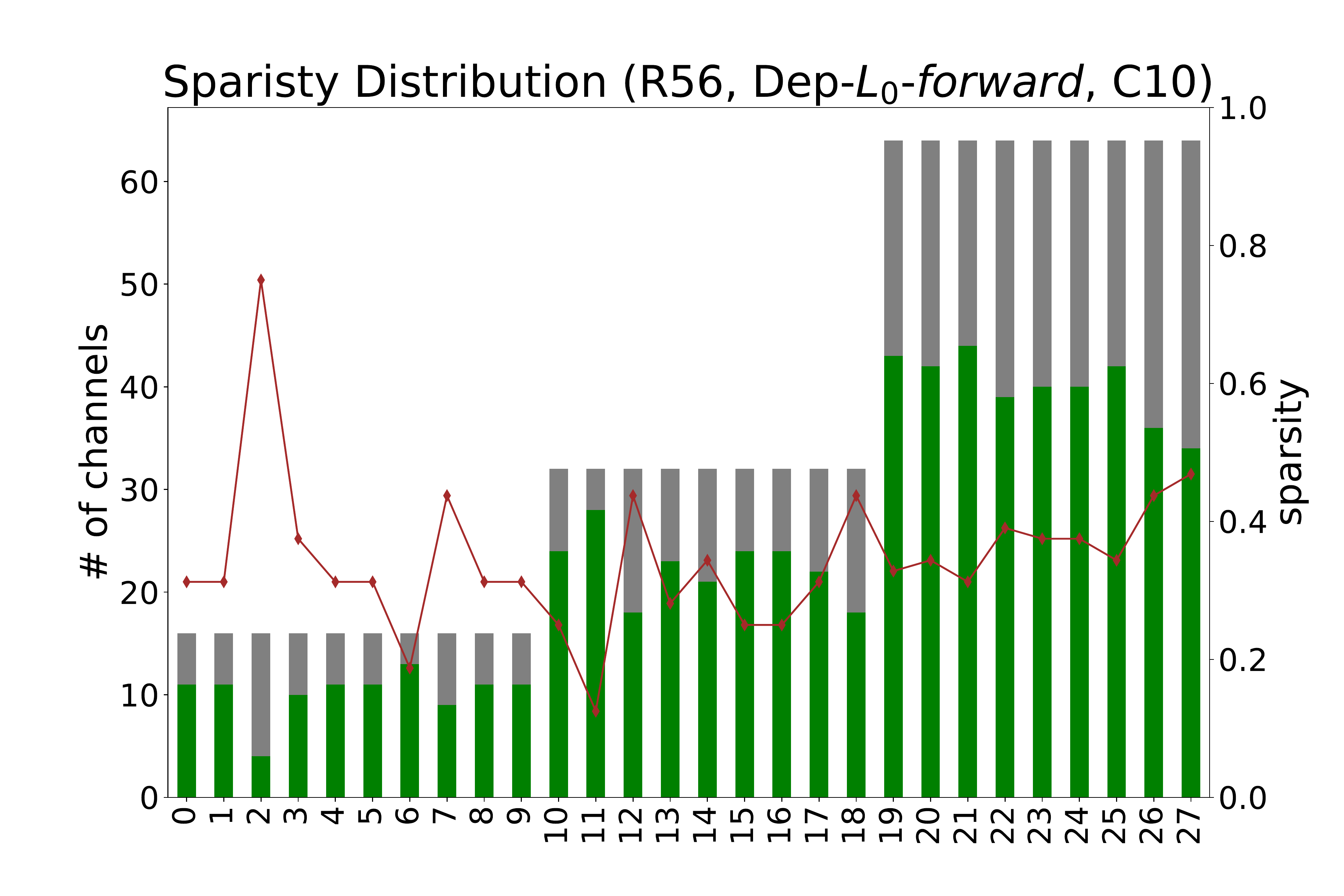}}
      \hfill
      \subfigure[R56-C10 ($L_0$-HC)]{\includegraphics[width=0.48\linewidth]{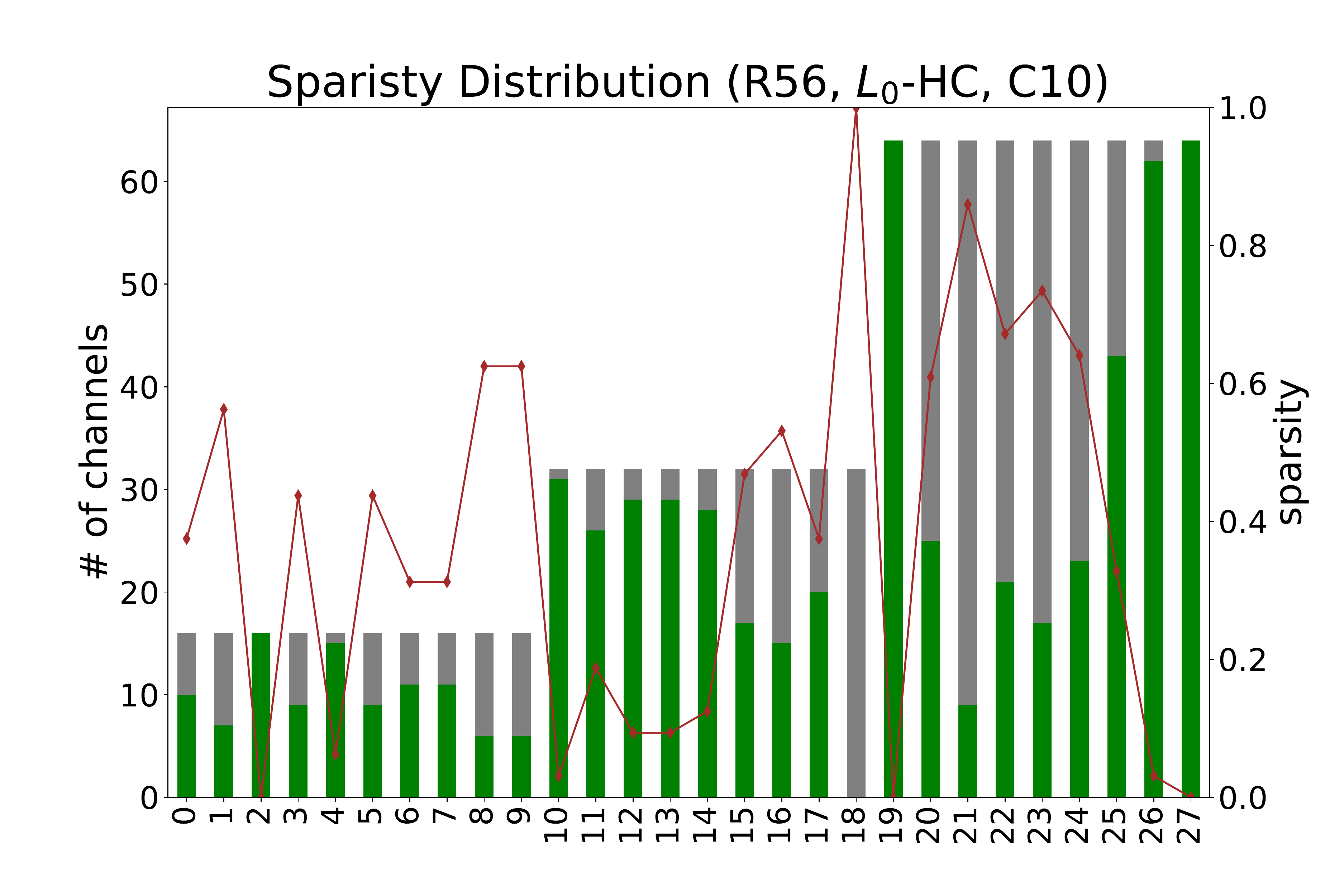}}
    \end{center}
\end{figure}

Fig.~\ref{fig: dist of sparsity2} shows the layer-wise prune rates of Dep-$L_0$ and $L_0$-HC when training ResNet56 on CIFAR10. Similar to the observations in the main text, Dep-$L_0$ sparsifies each layer of ResNet56 by a roughly similar prune rate, while $L_0$-HC tends to have dramatically different prune rates for different layers (sometimes prunes all or none), shedding some light on why $L_0$-HC has an inferior performance compared to Dep-$L_0$. 

\end{document}